**Design and Construction of Unmanned Ground Vehicles for Sub-Canopy Plant Phenotyping**

Authors: Adam Stager[1,2], Herbert G. Tanner[1], and Erin Sparks[3*]

1. Department of Mechanical Engineering, University of Delaware, Newark, DE 19711
2. TRIC Robotics LLC., Newark, DE 19711
3. Department of Plant and Soil Sciences, University of Delaware, Newark, DE 19711

*Corresponding Author: esparks@udel.edu

**Running Head**

Unmanned Sub-Canopy Plant Phenotyping

**Summary/Abstract**

Unmanned ground vehicles can capture a sub-canopy perspective for plant phenotyping, but their design and construction can be a challenge for scientists unfamiliar with robotics. Here we describe the necessary components and provide guidelines for designing and constructing an autonomous ground robot that can be used for plant phenotyping.

**1. Introduction**

Advances in genetics research have revolutionized the agricultural industry in ways that have vastly improved crop yield and resistance, however progress has slowed due to the challenges of identifying desirable genetic traits. Some beneficial traits are identifiable by physically measuring features of plants and can be measured non-destructively using portable handheld tools. While manual field measurement is possible, given that there can be thousands of plants per acre, and factoring in labor costs and variability between human crop scouts, consistent data acquisition at the scale necessary to integrate phenotyping into genetic pipelines can be difficult to obtain. Fully and semi-autonomous robotic systems offer a solution for gathering vast amounts of data in the field to relieve the phenotyping bottleneck [1].

The goal of this manuscript is to offer a guide, in layman's terms, on how to design and construct a ground mobile robot. It targets an audience of scientists or practitioners with little or no prior knowledge in robotics. We begin with a general discussion on component selection, to guide design choices for systems tailored for sub-canopy data collection. Then, a detailed example outlines the construction of a tracked robot platform (see Fig. 1) intended for sub-canopy terrain conditions in corn fields at the University of Delaware.

**2. Design Considerations**

Whether a robot is to be built from scratch or purchased off the shelf, there are many factors to consider when choosing the right equipment. By combining our own experience with that found in literature, we bring together important design considerations for ground mobile robots. We step through each component of a sub-canopy robot and provide some practical insights to streamline the design process.

2.1 Application Constraints

The scientist's crop and traits of interest determine the robot's specifications and set quantitative robotic design metrics (i.e. width must be less than 20 inches, sensor height must be greater than 8 feet) that help a robot designer compare tradeoffs between features of the robot. For example, the amount of money one can allocate to the development of a robot might put a constraint on cost. Because sub-canopy robots operate in a particularly confined environment, some of the important metrics relevant to sub-canopy phenotyping are discussed.

Minimum row spacing is one field characteristic that has an important impact on the size of the robot. The width of the robot is typically constrained to the planting row width minus two times the maximum stalk diameter (see Note 1). Ideally a robot should be as narrow as possible; a thin design can accommodate better motion uncertainty and reduce the chances of collision with plants. There are tradeoffs, however: decreased width will reduce stability in the roll axis (tipping to the left and right when driving forward down the row). If the terrain is expected to be very flat and dry, then the tip-over risk is less significant.

Crop height and intended robot and/or plant localization accuracy are factors that affect the robot's height and sensing capabilities. In general, it is best to keep the robot's center-of-mass low (see Note 2) in order to increase stability, however sub-canopy systems may be required to measure features at the plants extrema where the top canopy can exceed heights of 15 feet (e.g. bioenergy sorghum [2]). If measurements are required at a prescribed height along the stalk or information on average crop height is to be measured, a lightweight mast can provide a solution. This mast can double as a mount for an accurate global positioning sensors (GPS) known as a real-time kinematics (RTK) GPS but coupled with rough or uneven terrain a mast can also cause tipping, positional errors, and may interact with the top of the plants as shown in Fig. 2.

2.2 Frame

After setting the restrictions on the robot's physical dimensions and design configuration, the mechanical structure of the robot is considered. Some off-the-shelf designs offer pre-built frames (e.g. SuperDroid Robots, Inc.) that can be easily modified [3]. Generally, it is more practical to keep the frame lightweight because weight impacts overall runtime; in addition, lighter robots are easier to transport. A heavy robot can also leave depressions on the ground and is more likely to get stuck in mud. Cost often limits the use of exotic lightweight materials such as composites and titanium. Painted and galvanized steel or aluminum are common choices because they offer good structural rigidity at a relatively low cost. Although steel is heavy and more difficult to machine, it is easy to weld, and its low cost makes it a good choice for prototype designs. Aluminum is a convenient, lightweight material that can be bent into complex shapes, machined at low cost, while resisting corrosion. Some designs offer modularity by using a combination of clamps and tubes [4].

2.3 Drivetrain

The drivetrain is the system that allows the robot to get around in its environment. It can be configured in many different ways [5], and typically includes as major components wheels or tracks, motor (of different size and placement), motor drivers and controllers, and batteries (see Fig. 3). Sub-canopy robots are most often designed with either tracks or wheels. Tracks tend to cause less rutting and are less likely to get stuck in rough terrain, but they require more power, are more expensive, typically require more complicated transmission mechanisms, and can pick up weeds or tillers. Wheeled setups are often four-wheel drive, with one side running independently of the other (differential drive), allowing the robot to turn in place. Wheels, however, are susceptible to high centering –when the center of the robot gets stuck on rough terrain—but their low cost and ease of use makes them a good option when the terrain is not particularly challenging.

Motors come in several forms: there are brushed, brushless, stepper, and servo motors, but the easiest to control and most commonly used for ground mobile robots are brushed direct current (DC) motors. Other types of motors require special sensing and/or timing control to make them spin, whereas brushed motors contain physical brushes (hence their name) that connect electrically their spinning and stationary parts. The speed and torque requirements are two specifications that guide the selection of motor size. Knowing how much field coverage is desired is a good place to start because it can narrow motor and gearing combinations to a desired number of revolutions-per-minute (RPM). For example, the TERRA-MEPP sub-canopy platform used two 24VDC DG-158A wheelchair motors with a 135 RPM no-load output speed to cover 0.4 hectares in two hours [6]. Average motor torque is difficult to compute a-priori during the design process because calculations rely on complex ground interactions (depending on if it is muddy, dry, level of weeds or tiller interaction) and friction between transmission components, however it is possible to calculate upper bounds that help with motor selection.

Once the brushed DC motors are selected, they should be paired with a brushed DC motor driver or motor controller. These electrical components allow the robot's on-board central processing unit (CPU) to control the amount of power sent to the motors. A motor controller is typically more expensive than a motor driver, because it has additional functionality such as speed and position control (when encoder feedback is provided) or switching between remote control (RC) and autonomous control modes. The specification sheet associated with the motors will give information on "amperage at stall torque" which is an indication of how much the motor driver/controller should be able to output in the worst (most challenging) case. If a stall condition is encountered in the field, an undersized motor driver/controller can overheat causing thermal shutdown or damage to the circuitry (see Note 3).

Battery selection bridges the gap between drivetrain and electrical subsystem because it influences the entire system. Ultimately the size of the battery, described in Amp-hours (Ah), should be determined based on the estimated power draw from the motors plus the power draw of all sensors and the CPU. Most of these values can be found on specification sheets, but the motor output is heavily dependent on sizing, losses in power transmission, and environmental conditions. Similarly, with motor sizing, a bound on maximum and minimum runtimes of the robot and average working conditions can help establish an average battery power consumption, helping to determining proper battery capacity.

2.4 Electrical

It is unlikely that a robot will run on a common, consistent voltage, as different components may require different input voltages. Once the robot's function and capabilities are well established, a designer may assemble a wiring harness with carefully measured lengths and connectors that are mounted with zip ties or cable clips along the frame of the robot. During the prototyping phase however, it is useful to have flexibility so that new sensors can be tested without redesigning the entire electrical system. It is useful to include a power and ground distribution block, which is a physical component connecting wires to a single source, typically using convenient screw terminals. Voltage regulators can help adjust the voltage depending on the requirements of the motors, sensors, and CPU. Depending on expected operating conditions, moisture can be an important consideration affecting how to weather-proof the electrical components. A completely sealed box integrated into, or attached to, the frame can provide a waterproof setup but operating such a system in high temperature environments may cause overheating of the electronics. Off-the-shelf CPU coolers are available with incorporated air filters to prevent dust from entering the compartment and can be mounted to an electrical box to maintain safe operating temperatures (see Note 4).

2.5 On-board Computer (CPU)

An on-board computer is necessary if a designer wants to do any computation on the robot. Selection of an on-board computer depends heavily on both the desired level of autonomy and number of sensors that will be connected to the system. For cases where very little processing is required, a microcontroller such as an Arduino, which can only run a single script, may suffice. For navigation and real-time processing, however, a microprocessor is needed. The processing power of microprocessors found on single board computers, such as the Raspberry Pi 3, are generally limited but may still be suitable for applications with minimal to moderate autonomy or sensing. Fully-autonomous or semi-autonomous ground robots would likely require a mini PC. The Intel Nuc is one commonly used mini PC because of its highly compact form factor and range of available options for processing power. Scientists who require vast amounts of sensor data for processing feedback in real time have also networked multiple dedicated computers serving different subsystems to avoid processing conflicts between subsystems [7].

For navigation, some off-the-shelf autopilots such as PixHawk or the Novio2 can be combined with freely available software to streamline the problem of navigating the robot along a path, but are limited to common robot configurations and they are not as flexible for path planning as a custom setup. For systems where geolocation is not required (i.e. when tags in the field identify plants) or if the ground platform is to be teleoperated, then the on-board computer and sensing system can be decoupled from the motor control.

2.6 Navigation Sensors

Data from several sensors can be fused together to get accurate position and orientation of the robot as it travels through the field. If precision planting coordinates are available, then paths through the rows can be established, otherwise the robot can be driven manually while capturing GPS data to identify waypoints. If the robot is to navigate without relying on some assumed plant arrangement and/or centimeter-level accuracy is needed while following a predetermined path, then the accuracy of a real-time-kinematic (RTK) GPS is indispensable. For sub-canopy robots, the GPS is typically placed on top of a mast in order to obtain a good signal; the canopy will otherwise obstruct the connection and make GPS unreliable. When the GPS unit is attached to a vertical mast, then errors in position due to the roll and pitch of the robot can be significant, and naturally increase with mast length. An accelerometer (see Note 5) on the base can mitigate the problem by providing additional data for tilt correction. In conditions where wheel slippage is minimal, encoders can be used to get reasonable estimates of position, although one has to keep in mind that skid-steer vehicles rely on wheel slippage during turns, and that introduces errors in estimates of orientation. A gyroscope (see Note 6) coupled with a magnetometer (see Note 7) can be used in tandem to resolve unambiguously the platform's orientation. Fusing sensor data in a Kalman Filter (see Note 8), a standard robot position/velocity estimation algorithm, can help obtain better position estimates; Kalman filter implementations are available in several open-source software packages. Redundancy in sensor data becomes increasingly important as the robot's speed increases, especially considering the relatively slow 5-10Hz data rate from GPS, and can make position estimation more robust to occasional sensor outliers and failures. Using visual odometry (see Note 9), a state-of-the-art method of camera feeds with other sensor measurements and for which implementation software is also freely available, is another promising strategy for localizing a robot; it does require, however, relatively more processing power, careful calibration and can be prone to errors when used in highly dynamic environments.

2.7 Phenotyping Sensors

There are many imaging modalities available for collecting data relevant to phenotyping, however for in-situ field sensing, we consider only non-destructive samplers, cameras, and sensors (see Note 10) that can be readily mounted on a ground mobile robot. To help with the daunting task of selecting between the wide range of available options, this section briefly describes some of the most common sensor types and how they can be used for phenotyping.

2.7.1 RGB camera

Arguably the most common sensing modality, single-sensor cameras capturing visible light (VIS) can be used with no geographic data if some label is present on a plant to associate it to its images [8]. Images can be taken at high resolution at the expense of a larger file size. Many features can be identified from VIS data, especially in structured environments when software tools such as PlantCV (http://plantcv.danforthcenter.org/) are available [9]. However, differences in lighting and occlusions in the field may require additional processing for data analysis. Recently, machine learning algorithms have been used to extract features from plants by first training such algorithms with datasets labeled by human experts [10]. Some common features extracted include leaf area index (LAI), plant height, stem thickness, yield estimates, leaf and stand count [11, 12].

2.7.2 Spectral camera

Spectral cameras are designed to pick up light from individual spectra not visible to the human eye. By focusing light emission as a result of excitation by specific wavelengths of light, scientists can study internal characteristics of a plant non-destructively and before they become apparent by sensing visible light. The Normal Difference Vegetative Index (NDVI) and Photochemical Reflective Index (PRI) are two classical indices obtained by spectral imaging and used to quantify plant health [13, 14]. There are two categories of spectral cameras – hyperspectral and multi-spectral. Hyperspectral cameras are orders of magnitude higher resolution than multi-spectral cameras and are a recent focus of research. These cameras can capture thousands of narrow bandwidths compared to multi-spectral cameras which acquire data from 5-12 much wider spectral bands, however the data from hyperspectral cameras can be overwhelming and the requirement for careful calibration makes these new cameras impractical for some applications [15].

2.7.3 Stereo camera

By combining two cameras with a known separation distance, stereo cameras use correspondence between images to calculate distances in the form of disparity maps and provide estimates of depth for objects in the image. Accurate distance estimates rely on matching features between images and perform poorly in low-light where features can be difficult distinguish [16].

2.7.4 Time-of-flight (ToF) sensor

By emitting infrared (IR) light and measuring the return time of the reflected light on a low-resolution camera, distances can be estimated directly. These sensors are generally low resolution compared to RGB cameras but can offer depth without the computation of stereo camera setups. ToF cameras work well in low-light, but are susceptible to noise in direct sunlight because the sun's IR emission can conflict with the sensors output IR. Despite their relatively low resolution and sensitivity to ambient light, ToF are generally favorable for determining features in outdoor agricultural settings [17].

2.7.5 RGBD camera

By combining RGB with a ToF camera, RGBD cameras give a similar output as stereo cameras, but require less computation. Similar to ToF sensors, RGBD can be sensitive to lighting conditions and they have limited range due to the necessary matching between depth and RGB cameras. The Microsoft Kinect is an example of a low-cost RGBD camera that is commonly incorporated into robotics projects.

2.7.6 LIDAR

Light detection and ranging (LIDAR) uses a pulsed laser to measure distances at very high resolution with minimal noise compared to stereo and ToF camera sensors. Although these sensors can capture features in very fine detail, they require accurate knowledge of the position of the sensor and are significantly more expensive than other remote sensing methods. Sensor feedback is expressed in dense point clouds and can generate massive amounts of data that can be challenging to abstract to useful features [18].

2.8 Communication

Bandwidth and signal attenuation are the most important considerations when deciding on how to communicate between the robot and a ground station (a stationary computer where the human operator tracks robot progress). While an autonomous robot is capable of navigating on its own, keeping human operators in the loop safeguards against unanticipated challenges. For remotely controlled systems communication delays (latency) should be reduced because a lag in video or control command can result in crop damage.

Generally, commands are sent over radio frequencies; the range is exceptional, and real-time control commands usually require very low bandwidth. Wi-Fi is another way for communication with field robots offering higher bandwidth and allowing for video transmission, but at the expense of decreased range. For systems operating in range of cellular towers LTE can also be used to transmit high bandwidth data. Although cellular LTE offers superior range it can have fluctuating latency and is generally too slow for real-time teleoperation. Analog video has been gaining popularity for remote controlled vehicles, driving costs down and increasing reliability for these components. Although the analog video feed is not directly suitable for processing, it is a good option for teleoperation because it provides a long-range option with low latency and very little setup compared to other methods.

2.9 Software

Depending on the level of autonomy required, it can be relatively quick to get a robot running in the field. Sensors usually come with software for operating them on Windows, Mac, and Linux making it possible to collect data out of the box. At a minimum, a robot can be set up with teleoperated navigation and an RGB camera with on-board data storage. Although this requires intervention by a human operator, it can provide useful data from the hard-to-reach sub-canopy region and is a good place to start. If plants are tagged with visible identifiers then a human can quickly assign images to different plants and computer vision algorithms can achieve measurements of stalk thickness, lead-area-index (LAI), canopy density and other useful characteristics.

3. Construction of a Tracked Robot Platform

To provide a concrete example, we describe next the construction of a basic phenotyping robot. Off-the-self components are used wherever possible to make this platform reproducible with very little robotics background and at relatively low-cost. Specifically, the base is a SuperDroid LT2 Tracked ATR package including the frame, drivetrain, power and electrical components (see Note 11).

3.1 Materials

- Frame (SuperDroid LT2 Tracked ATR Robot Platform)
- Drivetrain (Tracks + Tensioner + Chain Offset)
- Power and Electrical (24VDC Lead Acid, RoboteQ MDC2460 2x60A 60V motor controller, RC Controller, RC receiver)

- On-board Computer (Raspberry Pi 3 with 16GB SD Micro Card)

- Phenotyping and Navigation Sensors (Camera and Camera Rails)

- Communication (Wi-Fi and Radio)

- Software (Linux Operating System)

- Ground control computer (Linux Operating System)

- Misc Parts (Mounting hardware, Foam weather proofing tape, 5lb mounting tape, USB AB cable, RS232 Connector)

3.2 Tools

- Phillips screw driver
- Adjustable wrench
- 1/8" Allen wrench
- 8mm Allen wrench
- Chain breaker
- Measuring tape
- ¼" box end wrench
- Wire cutter
- Crimping tool
- Power drill
- 3/8" drill bit
- 15/32" drill bit
- HDMI cable
- Monitor
- Keyboard

- Mouse

3.3 Mechanical Assembly

The mechanical assembly of a tracked robot requires accurate spacing between track wheels and a mechanism for tensioning the belt. Driving the wheels directly from the output of the motor's gearbox is not recommended and can lead to pre-mature failure of expensive motor components. The schematic (Fig. 4) shows a visual representation of the most important components throughout the robot assembly but purchase of a SuperDroid robot kit comes with an instruction manual including images. Here we summarize the steps to build the LT2 Tracked ATR robot – all components listed are included in the SuperDroid robot kit except for the camera rails and clamps which were purchased from a camera accessory company, SmallRig.

1. Using a Phillips screwdriver and adjustable wrench, bolt front axles (shafts for mounting the track wheels) to aluminum frame with eight #10-32 bolts and nuts.
2. Mount two drive motors, IG52-04 24VDC 285 RPM Gear Motors using a Phillips screw driver to tighten four M5 machine screws per motor.
3. Using the adjustable wrench, loosely mount track tensioning blocks using #6 hardware.
4. Slide rear axle into the tensioning blocks (mounting points that will be used to tension the tracks) and tighten two lock collars using 1/8" Allen wrench to prevent the axle from sliding left or right.
5. Slide #25 sprockets over motor shafts, but do not tighten their set screws until Step 11.
6. Install back-handle strap using Phillips screwdriver to tighten four tensioning screws.
7. Assemble two drive wheels with sprockets and two idler wheels using a Phillips screwdriver and ten #10 screws per wheel.

8. Slide wheel with sprocket on the front axle using a thrust bearing as a spacer on the inner and outer faces of the wheel followed by a lock collar tightened using a 1/8" Allen wrench.
9. Slide the motor forward toward the front wheel then wrap the #25 chain around both sprockets.
10. Using a chain breaker, remove excess chain and connect chain master link.
11. Slide motor sprocket along motor shaft so it is aligned with wheel sprocket then tighten motor sprocket set screw with a 1/8" Allen wrench.
12. Pull motor away from wheel until chain is tight and then tighten motor mounting screws using Phillips screwdriver.
13. Repeat Steps 8-12 for the other side.
14. Slide the tension blocks all the way forward then roll 2.75 inch molded spliceless tracks over the front and rear wheel on each side.
15. Tighten the four tension screws with the Phillips screwdriver at the rear of the robot evenly until the tracks are tight.
16. On each side, using a measuring tape, measure from the front axle center to the rear axle center to make sure both sides have equal separation. If the measurement on the right side is larger than the left, then use the Phillips screw driver to tighten the tension screws on the left side until they are equal and visa-versa.
17. Insert a ¼" box end wrench between the robot frame and the track to tighten the four screws that hold each tension block to the robot frame.

18. Cut ¼" wide strips from the foam weather proofing tape, peel off adhesive backing and attach tape around the entire perimeter of the top panel. This is used to cover the opening in the top of the frame.
19. Mount top panel using 16 Phillips head screws using Phillips screwdriver.
20. Drill four 3/8" holes 1"x1" from the corners of the top mounting plate then mount small rig camera rail clamps using an 8mm Allen wrench (see Note 12). The rail clamps allow the mounting rails to be adjusted depending on the application.

3.4 Electrical Assembly

The electrical assembly describes the connection of batteries, motor drivers, power regulators, and sensors. Here we use two 12VDC Lead Acid Batteries in series to create a 24VDC power source. This is fed directly to the motor driver and regulated down to 5VDC for a Raspberry Pi 3 (on-board computer) and 12VDC to power an on-board Wi-Fi router. All components in this section are included in the LT2 Tracked ATR package except for power and ground distribution blocks, 5VDC regulator, 12VDC regulator, Raspberry Pi 3, and Wi-Fi router. The regulators chosen should provide enough Amps to power the electronics connected to them. A more powerful Wi-Fi router will reduce latency and its signal can be improved by adding a long-range antenna on the ground control station, robot, or both. The camera is powered directly from a USB port on the Raspberry Pi 3. IG52-04 24VDC 285-RPM gear motors are connected to a RoboteQ MDC2460 2x60A 60V motor controller. Make sure to consider locations for mounting the electronics and routing the wiring - these aspects are often overlooked (see Note 13).

1. Install two 12V 8Ah Sealed Lead Acid Batteries, one on each side inside the aluminum LT2 chassis.

2. Slide threaded rods through the battery mounting tabs of the aluminum frame and tighten nuts over each end with an adjustable wrench (see Note 14).

3. Use mounting tape or screws to attached power and ground distribution blocks inside the frame.

4. Drill a 15/32" hole at the rear of the robot (in black plastic switch plate) and slide power switch through the hole then tighten the nut, securing the switch using an adjustable wrench (see Note 15).

5. Use mounting tape or screws to mount power regulators (5VDC, 12VDC) inside the frame.

6. Use mounting tape or screws to mount RoboteQ MDC2460 2x60A 60V Motor Controller inside frame.

7. Connect the RoboteQ motor controller to USB port of the Raspberry Pi 3 using a USB AB cable.

8. Use RS232 connector to connect RC control to the RoboteQ.

9. Use camera rail clamp to mount camera in desired location on camera rails, then feed cameras USB cable back to a USB port on the Raspberry Pi 3.

10. Confirm connections with the schematic in Fig. 5 before powering on the robot.

3.5 Software Setup

Software is an important part of a robot and can be increasingly complex depending on the level of autonomy desired. On the other hand, without adding software, a teleoperated robot can still collect useful data. A bare-bones system where a GoPro or other standalone camera can be attached to the robot as it is driven through the field, images can be extracted afterward by retrieving them from the camera's memory card. This method is very limited because the robot has no control over the camera or it's data. Instead we offer a middle ground solution by describing a setup that can be easily expanded upon. A single camera onboard the robot streams video over Wi-Fi to the ground control station where the user can collect images in real time while also using the camera to navigate the robot using RC control. Additional cameras or sensors can be added, and the control can be automated.

The robot uses the open-source Robot Operating System (ROS) on a Linux Ubuntu Mate operating system and sends data to a ground control computer running Linux Ubuntu 16.04 LTS. We assume Linux is installed on both ground control station and on-board computers (see Note 16). ROS is a set of software utilities and libraries that is convenient because it allows for easy implementation of new sensors and stores/handles sensor data in a convenient organized structure. It's worth mentioning that these steps can be challenging to approach with no prior experience with Linux, Raspberry Pi 3, or ROS. We hope that the steps provided will help to guide an inexperienced reader and have provided links where necessary to help reinforce more complex steps.

1. Startup (boot) Linux (Ubuntu Mate distribution) by powering on the Raspberry Pi 3 (see Note 17).
2. Install ROS Kinetic using the command line (see Note 18).
    a. Boot Raspberry Pi 3 with Ubuntu Mate installed.
    b. Connect Wi-Fi to internet, open a web browser and navigate to http://wiki.ros.org/kinetic/Installation/Ubuntu where line by line commands can be copy/pasted into the terminal (where command line text can be run).

c. Highlight each command and right click on the computer mouse to select copy, then right click in the terminal window and select paste.

   d. Press Enter to run each command.

   e. Once complete, test the installation by first navigating to http://wiki.ros.org/ROS/Tutorials/InstallingandConfiguringROSEnvironment in the web browser.

   f. Create a ROS workspace by copy/pasting terminal commands as in Steps 2.c-2.d.

3. Download the usb_cam ROS node (http://wiki.ros.org/usb_cam) which is a collection of programs and camera utilities packaged with the camera driver and various parameters that make it easy to use. Calibration is one file that should be updated depending on the camera that is used. In this step it is easiest to plug in an HDMI cable, mouse and keyboard to confirm the camera works directly on the Raspberry Pi 3.

   a. Once the ROS environment is created open a new terminal.

   b. Install the usb_cam ROS package by typing "sudo apt-get install ros-kinetic-usb-cam" followed by Enter

   c. Once installed plug in the USB camera and open two separate terminal windows. In the first window run "roscore". In the second terminal window run "rosrun usb_cam usb_cam_node". This will publish the raw data from the camera to the computer.

   d. To view the camera data open one more terminal window and run "rqt_image_view". From the pull down in the top left of the popup window select "/usb_cam/image_raw" to confirm the camera output.

4. Setup remote camera triggering and video streaming over Wi-Fi.

a. Plug the Raspberry Pi 3 with Ubuntu Mate and ROS into the robot's Wi-Fi router using an ethernet cable.

b. Boot the Raspberry Pi 3 while connected to HDMI and type "ifconfig" into a new terminal. Find and take note of the "inet addr" which will start with 192.168.1.x where x is a unique identifier designated to the Raspberry Pi 3. Now the HDMI, keyboard, and mouse can be removed from the Raspberry Pi 3.

c. On the ground control computer log into the robot's Wi-Fi network.

d. On the ground control computer open a new terminal and run "ifconfig" to identify the ground station's "inet addr". It will start with 192.168.1.y where y is a unique identifier designated to the ground control computer.

e. Log into the Raspberry Pi 3 remotely using secure shell (SSH) (see Note 19).

5. Stream live video from Raspberry Pi 3.

    a. Complete Step 4 making sure to note the unique addresses 192.168.1.x and 192.168.1.y where x and y are numbers unique to the Raspberry Pi 3 and ground control computer respectively.

    b. Boot the Raspberry Pi 3 and power the robot's Wi-Fi router (see Note 20).

    c. Open three terminal windows on the ground control station's computer.

    d. In the first terminal, log into the Raspberry Pi 3 remotely as in Step 4.e. (see Note 21).

    e. In the second terminal run "roscore"

    f. Return to the first terminal and run "rosrun usb_cam usb_cam_node"

g. In the third terminal run "rqt_image_view". From the pull-down in the top left of the popup window select "/usb_cam/image_raw". The robots video feed should be displayed.

h. Save images by clicking on the save file icon to the far right of the pull-down selection. This is not a fast way to collect images from the robot but will give a feel for what it is like to teleoperate a system and get data remotely. It will help with understanding the challenges of latency and can be extended to autonomously capture images or include multiple camera.

**4. Notes**

1. It may be more relevant to consider an estimated maximum stalk (base) diameter because branches, tillers, or brace roots extending into the row can disturb the robot's motion if they are not considered.
2. It is best to keep heavier components closer to the ground.
3. For four-motor configurations, the 4WD operation will be severely limited if current on each side of the robot is split between a single motor driver. In this case, the designated motor input will travel the path of least resistance and insufficient traction for either wheel on a particular side will result in wheel spinning and loss of forward motion. It is best to allocate either one motor driver per motor or one dual motor driver per side to get true 4WD traction.
4. It is practical to pair the motor voltage with the battery output because motor output will be a heavy drain on the battery capacity. If the voltages do not match then a voltage regulator can change the battery output at the expense of a loss in efficiency during the voltage conversion. This would get you less motor output for the same battery charge and voltage regulators can also limit the output current to the motors.

5. Typically, accelerometers measure acceleration along their local x, y, z axes. By measuring the force of gravity an accelerometer can help inform the robot which direction is down. With additional computation it can also help estimate robot's position.

6. A gyroscope (digital) is used to determine orientation in roll, pitch, and yaw.

7. A magnetometer measures magnetic forces and acts as a compass for the robot. Magnetic fields from the motors can affect magnetometer readings so they should not be placed close together.

8. A Kalman Filter combines measurements from multiple sensors to reduce uncertainty.

9. Visual odometry is a method of determining position and orientation based on analyzing changes between camera images.

10. Cameras are also a type of sensor, but we typically see the output in the form of an image. The raw data is similar to other sensors. For example, an array of 10x10 individual photo (light) absorbing semiconductors would be a 100-pixel camera sensor. Digital values for red, green, and blue are captured for each pixel and then processed into the color image typically viewed.

11. The LT2 Tracked ATR package can also be purchased pre-assembled.

12. Four SmallRig 15mm camera rails, four SmallRig quick release clamps, four SmallRig 90-degree rod clamps.

13. We have found it useful to use 5lb outdoor mounting tape to mount electronics semi-permanently for when aspects of the robot are still in development. This prevents unnecessarily drilling extra holes in the robot's frame and saves time, but electronics should be mounted more permanently once the robot is complete. Alternatively, a mounting panel can be made with many holes drilled into it for mounting small parts. This way the mounting panel can be replaced when components change and as a result less holes are required in the robot's frame. The goal is to keep wiring neat and to keep wires short if possible. Zip ties can be helpful for securing wires along the frame. We leave the selected method of mounting the electronics up to the reader.

14. These rods act as battery tie downs to make sure the batteries don't shift into the CPU or other electronics.

15. A fuse that prevents too much current from flowing through the system can be added but is not required.

16. These operating system (OS) procedures are well documented and outside the scope of this work.

17. In order to install Ubuntu Mate on a Raspberry Pi 3 a 16 GB (minimum) SD micro memory card is needed. If it must be installed then: download Ubuntu Mate for free (https://ubuntu-mate.org/download/), format the SD card, then using free Win32DiskImager (https://sourceforge.net/projects/win32diskimager/) the .img file for Ubuntu Mate can be added to the SD card. Insert the SD into the Raspberry Pi 3 and follow the prompts to set up Ubuntu Mate. If the ground control station's computer does not have Linux then a "bootable USB" can be created using a .iso image of Linux distribution Ubuntu 16.04 LTS. Booting the computer from the USB will guide setup alongside windows or as a standalone operating system. These Linux distributions were chosen because they have been shown to cause the fewest difficulties in installation and use, but they can be switched with newer versions as long as they support ROS.

18. The command line is a text interface where you can launch commands (run programs) by typing specific text into the interface window, called the terminal. You can open a terminal by pressing CTRL+ATL+T.

19. SSH is a way of sending commands to a computer on the same network. Knowing the login name and password to the Raspberry Pi 3, log into it from the ground control station computer by opening a new terminal and running the command "ssh login@address". Where "login" is the login name of the Raspberry Pi 3 and "address" is the inet address found by running "ifconfig" in Step 4.b. A prompt will request the password from the Raspberry Pi 3. Now this terminal window on the ground control computer can be used to run programs on the robot remotely.

20. If the Raspberry Pi 3 and Wi-Fi router are connected to the robot's power source, then flipping the power switch on the robot will power both on. Otherwise if they are not yet installed in the robot, power them independently.

21. If the Raspberry Pi 3's inet address is not recognized, then make sure the ground control station's computer is connected to the robot's Wi-Fi network. Sometimes when it is powered off it will default to a previously saved network and lose connection with the robot.

**Acknowledgement**

This work is supported by the Delaware Biosciences Center for Advanced Technology Entrepreneurial Proof of Concept Grant.

**References**

1. Furbank, R. T., & Tester, M. (2011). Phenomics - technologies to relieve the phenotyping

bottleneck. *Trends in Plant Science*, *16*(12), 635–644.

https://doi.org/10.1016/j.tplants.2011.09.005

2. Baharav, T., Bariya, M., & Zakhor, A. (2017). In Situ Height and Width Estimation of Sorghum Plants from 2.5d Infrared Images. *Electronic Imaging*, (17), 122–135. https://doi.org/10.2352/issn.2470-1173.2017.17.coimg-435

3. Zhang, J., Singh, A., Lofquist, A., Singh, A., Bhattacharya, S., Gao, T., … Sarkar, S. (2018). A Novel Multirobot System for Plant Phenotyping. *Robotics*, *7*(4), 61. https://doi.org/10.3390/robotics7040061

4. Grimstad, L., & From, P. (2017). The Thorvald II Agricultural Robotic System. *Robotics*, *6*(4), 24.

5. Bawden, O., Ball, D., Kulk, J., Perez, T., & Russell, R. (2014). A lightweight, modular robotic vehicle for the sustainable intensification of agriculture. *Australasian Conference on Robotics and Automation*.

6. Young, S. N., Kayacan, E., & Peschel, J. M. (2018). Design and field evaluation of a ground robot for high-throughput phenotyping of energy sorghum. *Precision Agriculture*. https://doi.org/10.1007/s11119-018-9601-6

7. Mueller-Sim, T., Jenkins, M., Abel, J., & Kantor, G. (2017). The Robotanist: A ground-based agricultural robot for high-throughput crop phenotyping. *Proceedings - IEEE International Conference on Robotics and Automation*, 3634–3639. https://doi.org/10.1109/ICRA.2017.7989418

8. Fahlgren, N., Gehan, M. A., & Baxter, I. (2015). Lights, camera, action: High-throughput plant phenotyping is ready for a close-up. *Current Opinion in Plant Biology*, *24*, 93–99. https://doi.org/10.1016/j.pbi.2015.02.006

9. Fahlgren, N., Feldman, M., Gehan, M. A., Wilson, M. S., Shyu, C., Bryant, D. W., … Baxter, I. (2015). A versatile phenotyping system and analytics platform reveals diverse temporal responses to water availability in Setaria. *Molecular Plant*, *8*(10), 1520–1535. https://doi.org/10.1016/j.molp.2015.06.005

10. Singh, A., Ganapathysubramanian, B., Singh, A. K., & Sarkar, S. (2016). Machine Learning for High-Throughput Stress Phenotyping in Plants. *Trends in Plant Science*, *21*(2), 110–124. https://doi.org/10.1016/j.tplants.2015.10.015

11. Li, L., Zhang, Q., & Huang, D. (2014). A review of imaging techniques for plant phenotyping. *Sensors (Switzerland)*, *14*(11), 20078–20111. https://doi.org/10.3390/s141120078

12. Mulla, D. J. (2013). Twenty five years of remote sensing in precision agriculture: Key advances and remaining knowledge gaps. *Biosystems Engineering*, *114*(4), 358–371. https://doi.org/10.1016/j.biosystemseng.2012.08.009

13. Gamon, J. A., Peñuelas, J., & Field, C. B. (1992). A narrow-waveband spectral index that tracks diurnal changes in photosynthetic efficiency. *Remote Sensing of Environment*, *41*(1), 35–44. https://doi.org/10.1016/0034-4257(92)90059-S

14. Tucker, C. J. (1979). Red and photographic infrared linear combinations for monitoring vegetation. *Remote Sensing of Environment*, *8*(2), 127–150. https://doi.org/10.1016/0034-4257(79)90013-0

15. Adão, T., Hruška, J., Pádua, L., Bessa, J., Peres, E., Morais, R., & Sousa, J. J. (2017). Hyperspectral imaging: A review on UAV-based sensors, data processing and applications for agriculture and forestry. *Remote Sensing*, *9*(11).

16. Kise, M., Zhang, Q., & Rovira Más, F. (2005). A stereovision-based crop row detection

method for tractor-automated guidance. *Biosystems Engineering*, *90*(4), 357–367. https://doi.org/10.1016/j.biosystemseng.2004.12.008

17. Klose, R., Penlington, J., & Ruckelshausen, A. (2011). Usability of 3D time-of-flight cameras for automatic plant phenotyping. *Bornimer Agrartechnische Berichte*, *69*, 93–105.

18. Omasa, K., Hosoi, F., & Konishi, A. (2007). 3D lidar imaging for detecting and understanding plant responses and canopy structure. *Journal of Experimental Botany*, *58*(4), 881–898. https://doi.org/10.1093/jxb/erl142

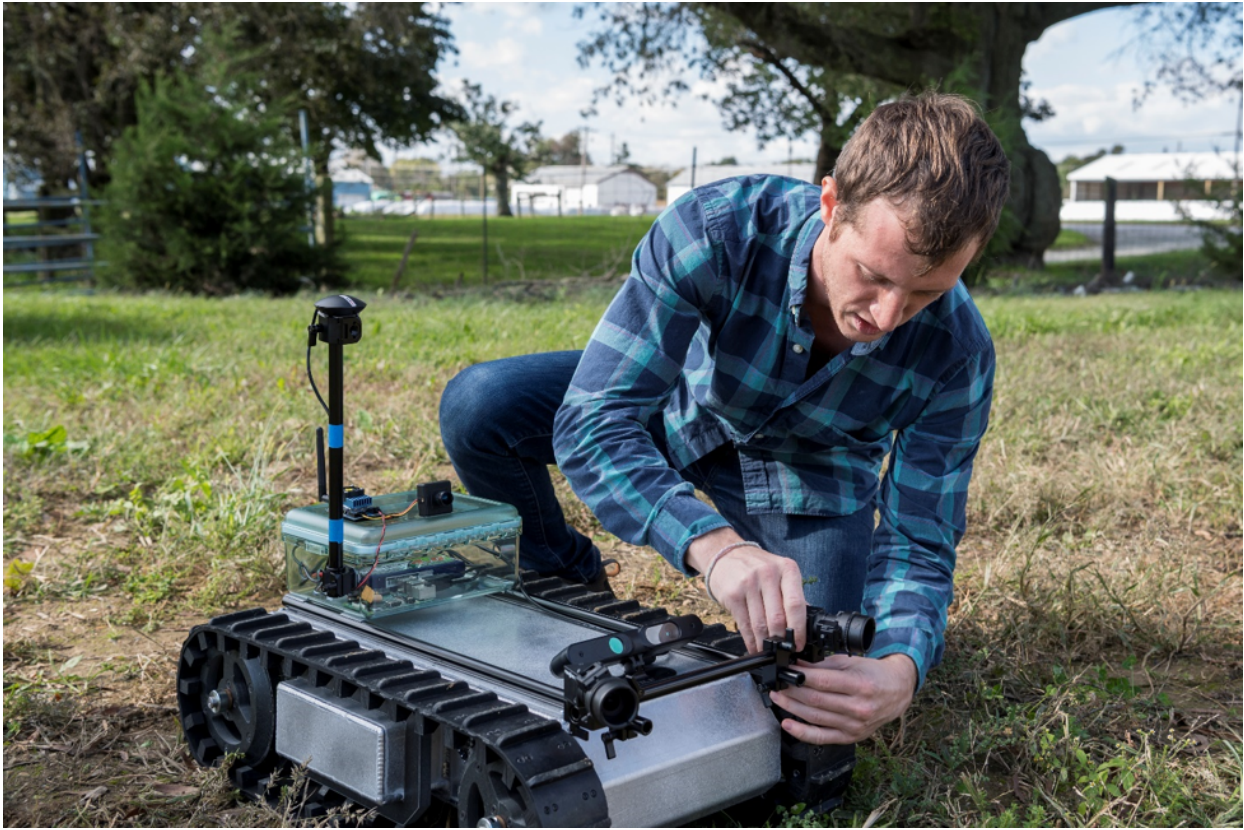

Figure 1. Adjusting cameras on a modified SuperDroid LT2 tracked ATR platform. For simplicity, the design described in this method includes a single camera sensor, but it is easy to add more sensing capabilities. Pictured here are additional sensors, including one extra RGB camera, an RGBD camera, analog video transmission, GPS, and external waterproof electronics housing. Camera rails allow quick and easy outfitting of the robot for various phenotyping experiments.

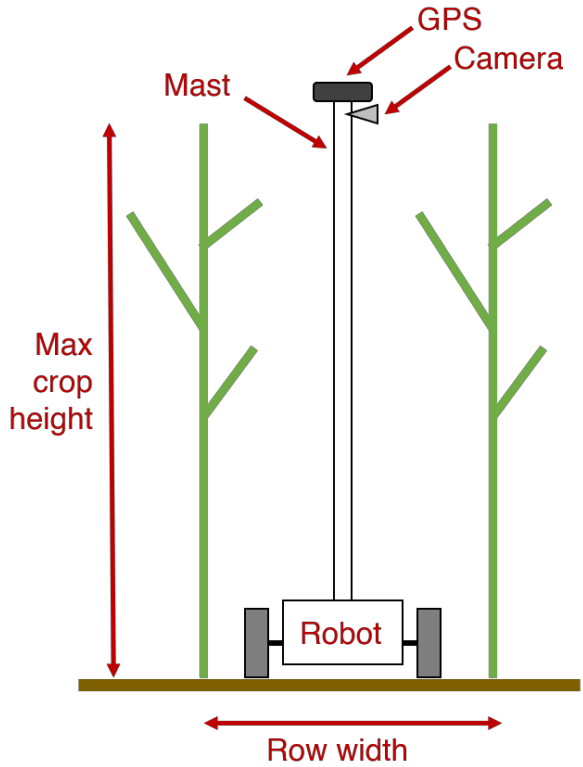
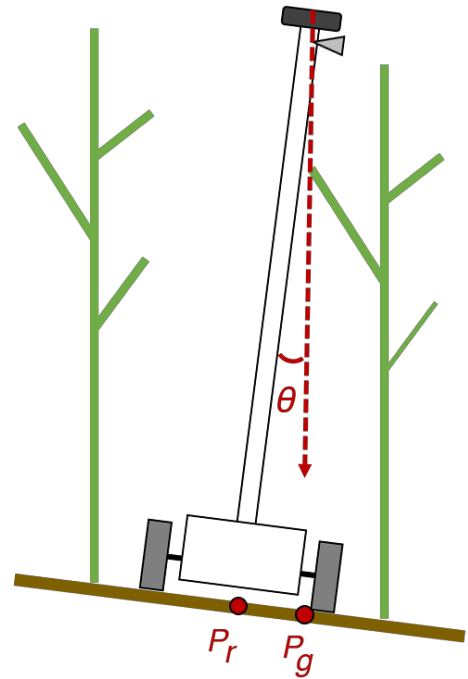

(a) Field dimensions can help determine robot constraints

(b) Tipping angle and incorrect position estimate

Figure 2. (a) Row width and maximum estimated crop height are important constraints for sub-canopy robot platforms, (b) Tipping angle $\theta$, caused by uneven ground or rough terrain can cause instability of the robot and lead to a false position estimate $P_g$ due to a resulting offset of the GPS from the robot's actual position $P_r$.

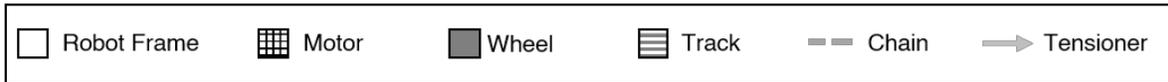

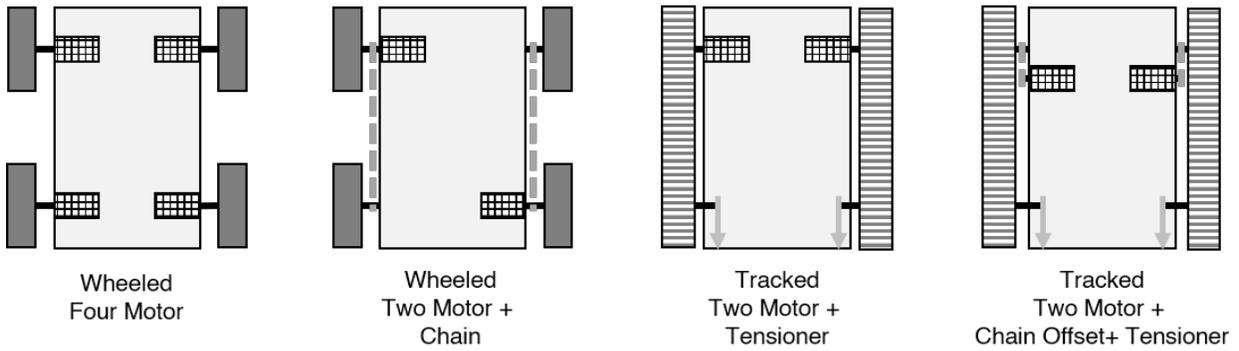

Figure 3. Examples of the most common configurations for ground mobile robots used for sub-canopy phenotyping. The arrows indicate a tensioner that moves in the direction of the arrowhead when it is tightened.

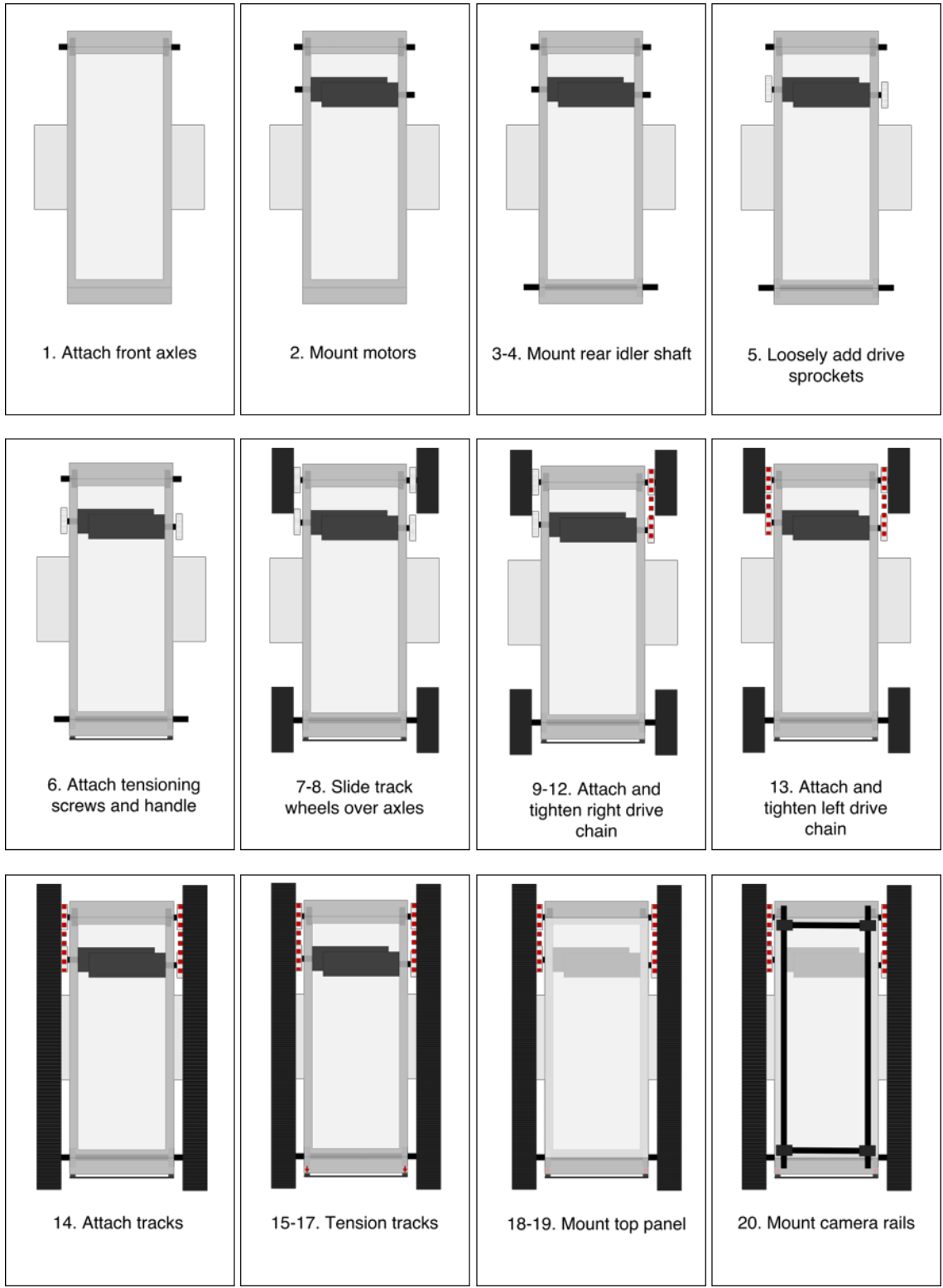

Figure 4. A visual guide representing the mechanical assembly of a SuperDroid LT2 tracked platform.

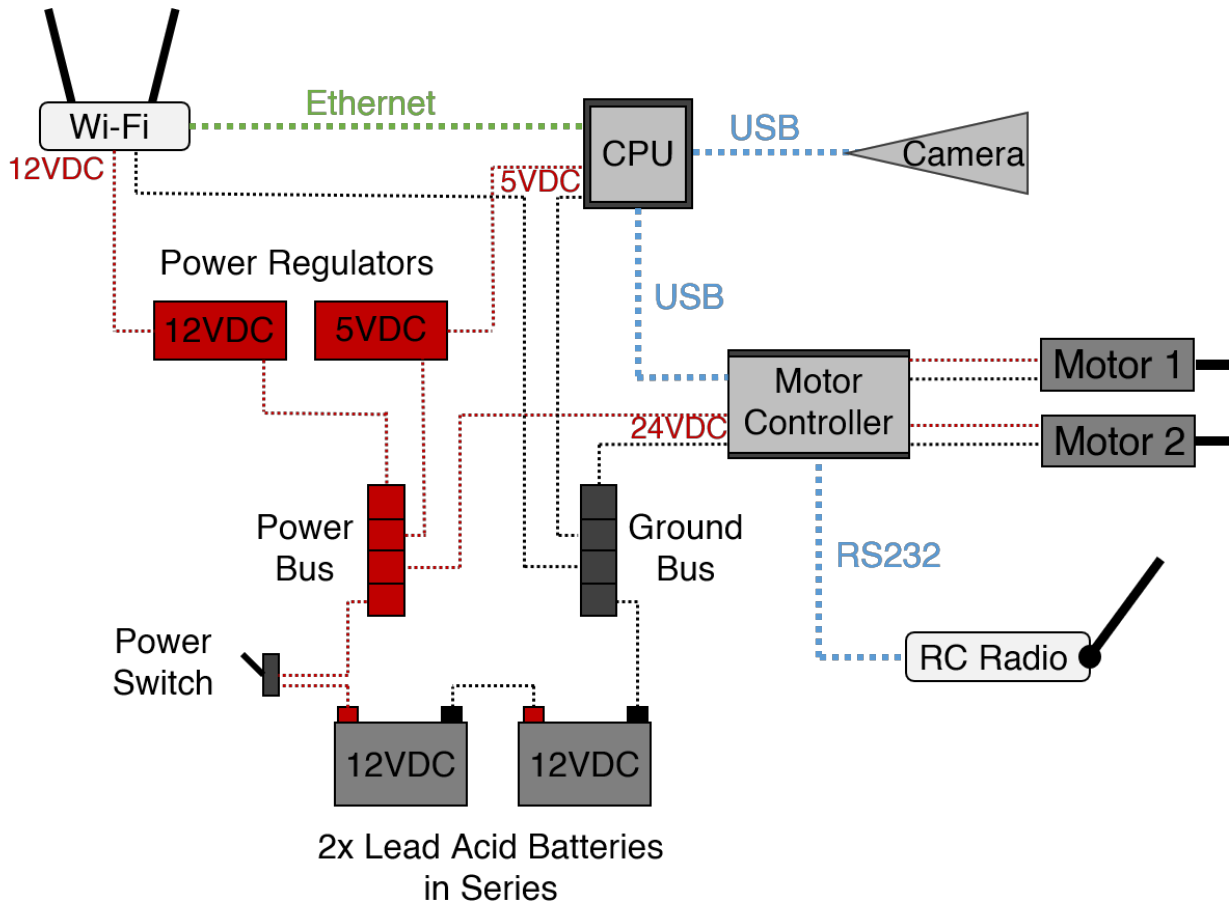

Figure 5. Schematic representation of electrical connections.